\documentclass{article}

\usepackage[english]{babel}

\usepackage[letterpaper,top=2cm,bottom=2cm,left=3cm,right=3cm,marginparwidth=1.75cm]{geometry}

\usepackage{amsmath}
\usepackage{amssymb}
\usepackage{graphicx}
\usepackage{booktabs}
\usepackage{comment}

\usepackage[pagebackref,breaklinks,colorlinks]{hyperref}

\newcommand{\bm}{\boldsymbol}
\newcommand{\ten}[1]{ \boldsymbol{\mathcal #1}}
\newcommand{\bbR}[1]{\mathbb{R}^{#1}}
\newcommand{\etal}{{\em et al.}}
\newcommand{\eg}{{\em e.g.}}
\newcommand{\ie}{{\em i.e.}}

\usepackage[capitalize]{cleveref}
\crefname{section}{Sec.}{Secs.}
\Crefname{section}{Section}{Sections}
\Crefname{table}{Table}{Tables}
\crefname{table}{Tab.}{Tabs.}

\usepackage{lipsum}
\newcommand\blfootnote[1]{%
\begingroup
\renewcommand\thefootnote{}\footnote{#1}%
\addtocounter{footnote}{-1}%
\endgroup
}

\title{{\Large Manifold Modeling in Quotient Space: \\ Learning An Invariant Mapping with Decodability of Image Patches}}
\author{Tatsuya Yokota$^{*,\star}$ and Hidekata Hontani$^*$ \\
  {\small $^*$Nagoya Institute of Technology, Aichi, Japan} \\
{\small $^\star$RIKEN Center for Advanced Intelligence Project, Tokyo, Japan}}

\date{ }

\begin{document}
\maketitle

\begin{abstract}
This study proposes a framework for manifold learning of image patches using the concept of equivalence classes: manifold modeling in quotient space (MMQS).
In MMQS, we do not consider a set of local patches of the image as it is, but rather the set of their canonical patches obtained by introducing the concept of equivalence classes and performing manifold learning on their canonical patches.
Canonical patches represent equivalence classes, and their auto-encoder constructs a manifold in the quotient space.
Based on this framework, we produce a novel manifold-based image model by introducing rotation-flip-equivalence relations.
In addition, we formulate an image reconstruction problem by fitting the proposed image model to a corrupted observed image and derive an algorithm to solve it.
Our experiments show that the proposed image model is effective for various self-supervised image reconstruction tasks, such as image inpainting, deblurring, super-resolution, and denoising.
\blfootnote{This work was supported by Japan Science and Technology Agency (JST) ACT-I under Grant JPMJPR18UU.}
\end{abstract}

\section{Introduction}
\label{sec:intro}
The non-local similarity (or long-range dependencies) in an image plays a vital role in various vision applications, and the similarity is generally measured by the relations between image patches \cite{buades2005non,zhang2019self,wang2018non}.
Non-local relationships in an image are exactly described by their affinity matrix.
The affinity matrix is a matrix whose size is the number of pixels by the number of pixels, and the $(i, j)$-th element of the matrix describes the strength of the connection between the $i$-th and $j$-th pixels.
A non-local filter (\eg NLM \cite{buades2005non} and BM3D \cite{dabov2007color,dabov2007image}) is a typical framework for image denoising that utilizes the non-local similarity of the images.
The (sparse) affinity matrix is usually obtained by the template matching between the image patches.
Self-attention structures (\eg a non-local block \cite{wang2018non} and self-attention modules \cite{zhang2019self}) used in deep learning are closely related to the non-local filtering with an affinity matrix.
The self-attention map corresponds to the affinity matrix, and it is usually calculated by the inner product of the channel fibers in feature maps, where each channel fiber of the feature map corresponds to the linear transform of each image patch in the case of convolutional neural networks (CNN).

%%%
The image prior hidden in the CNN structure has attracted attention due to the discovery of the deep image prior (DIP) \cite{ulyanov2018deep,ulyanov2020deep} in recent years, and its relation to nonlocal filters has been discussed by Tachella \etal \cite{tachella2021neural}.
In \cite{tachella2021neural}, Tachella \etal showed that the weight update in the CNN denoiser could be interpreted as twicing \cite{milanfar2012tour} (a non-local filter with feedback) in signal processing from the viewpoint of a neural tangent kernel \cite{jacot2018neural}. 

%%%%
The existence of similar patches in an image implies the low dimensionality of the manifold of image patches (\eg, manifold model \cite{peyre2009manifold,osher2017low}, patch-based denoising auto-encoder (AE) \cite{yokota2020manifold} ).
The latent variables encoded from image patches should be useful for the stable evaluation of the non-local similarity in an image.
However, local appearances are often degraded by nuisance operations such as noise addition.
The latent variables should be invariant against nuisance operations in order for the evaluation to be stable against degradation as the invariance would improve the quality of the image reconstruction.

\begin{figure}[t]
  \centering
  \includegraphics[width=0.6\textwidth]{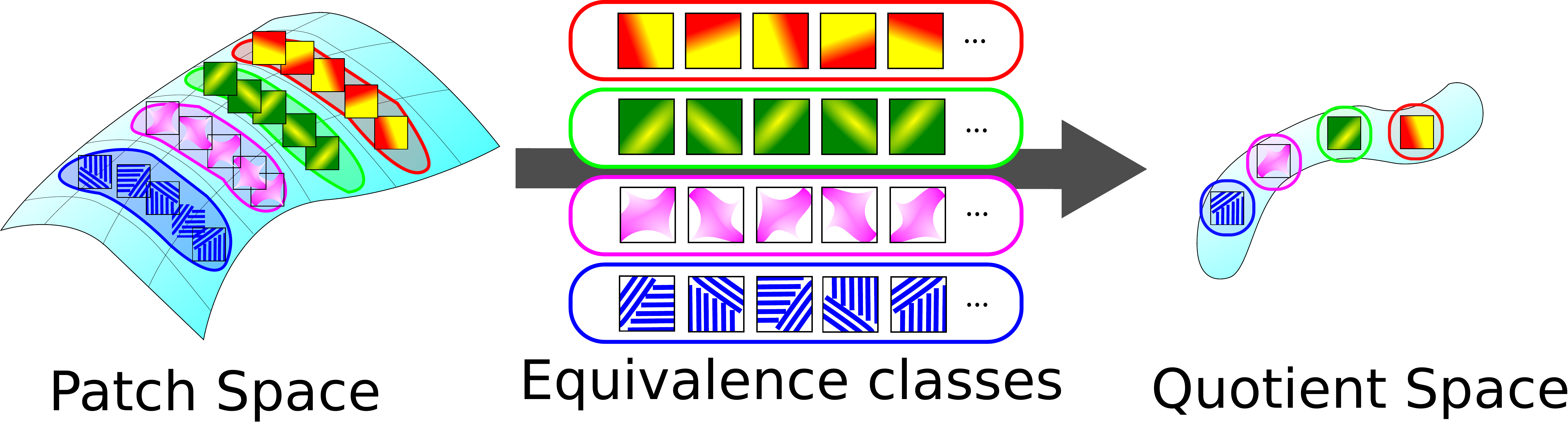}
  \caption{Image of a quotient space. The equivalence class is a subset of patches that have equivalence relations. The quotient space is a set of equivalence classes.}\label{fig:quotient_space}
\end{figure}

\begin{figure}[t]
  \centering
  \includegraphics[width=0.9\textwidth]{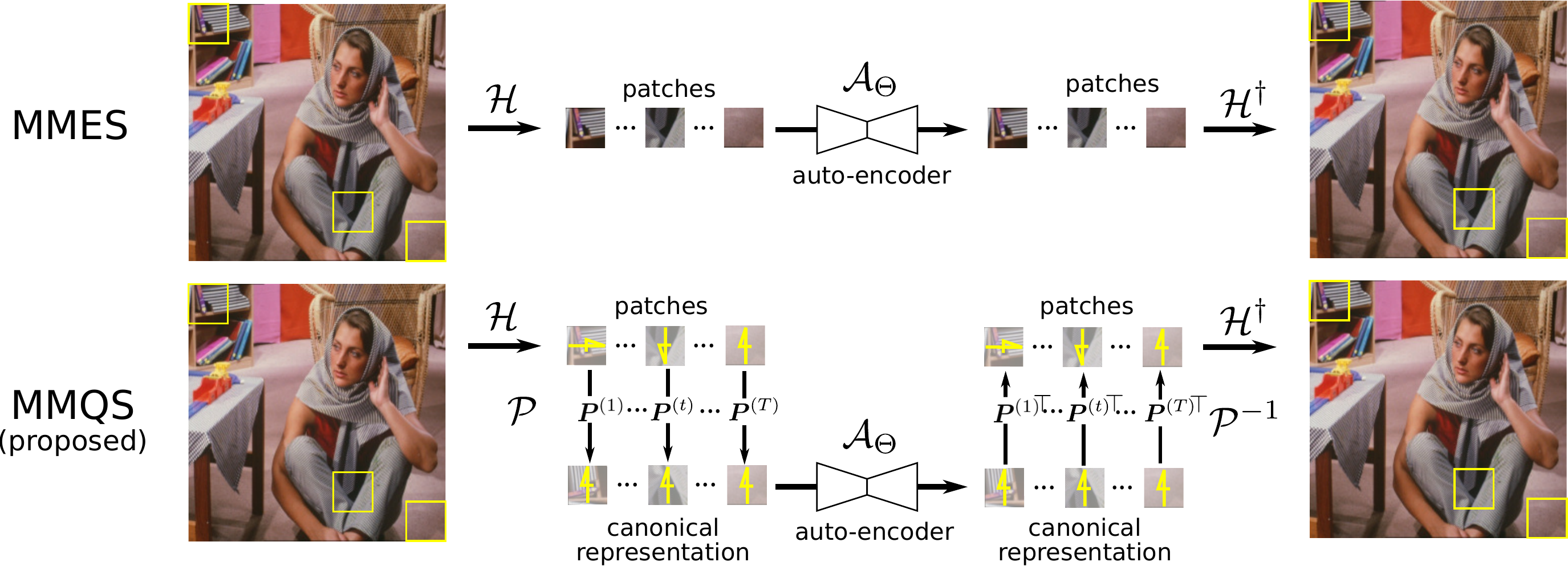}
  \caption{Structure of the proposed MMQS.  It consists of five parts: patch extraction $\mathcal{H}$, canonicalization $\mathcal{P}$, auto-encoder $\mathcal{A}_\Theta$, de-canonicalization $\mathcal{P}^{-1}$, and patch aggregation $\mathcal{H}^\dagger$. }\label{fig:concept_MMQS}
\end{figure}

In the image patches, we can find groups where the included image patches can be obtained by applying a geometric transformation to the template (the same image patch).
For example, image patches sampled along the smooth contour of an object in an image are obtained by rotating one of the sampled image patches.
Image patches sampled from a mirror-symmetric object can then be transformed to each other via flipping.
The long-distance similarity evaluated by the latent variables, which are invariant against geometric transformations, would help represent global image features, such as rotation- or mirror- symmetric structures.

However, the latent variables that are invariant against geometric transformations have no decodability of images, \ie, the given images cannot be restored from the latent variables.
This is because all image patches in the same group are encoded at the same point in the latent space (degeneracy).
To realize decodability, the degeneracy of the encoder should be avoided.
Thus, the image patches in each equivalence class are represented by a pair of invariant latent variables and their corresponding geometric transformation parameters.
The dimensionality of the resultant latent variables is significantly reduced because the patches in each class are encoded by the same latent variable. This dimensionality reduction significantly improves the performance of the vision tasks, especially when the number of image patches is limited or when some portions of image patches are lost or degraded.

In this study, we addressed the incorporation of invariance into patch manifold learning \cite{yokota2020manifold} by applying the concepts of rotation-flip-equivalence classes and quotient spaces.
By introducing equivalence classes as subsets of image patches that have equivalent relations (such as rotation-flip equivalence), a quotient space is defined as a set of equivalence classes (see \cref{fig:quotient_space}).
We can see that the latent variables that represent equivalence classes in a quotient space are invariant with respect to the rotation-flip manipulation in the original image space.
Subsequently, manifold learning is performed in the quotient space, hence the name: manifold modeling in quotient space (MMQS).
\cref{fig:concept_MMQS} shows the structure of the proposed MMQS and compares it to manifold modeling in embedded space (MMES) \cite{yokota2020manifold}.
MMES is based on three stages: delay embedding (patch extraction), patch AE, and inverse delay-embedding (patch aggregation).
In contrast, canonicalization and de-canonicalization layers are added before and after the patch AE in the proposed MMQS.
In the canonicalization layer, some rotation or flip operations are applied to each input patch to obtain a canonical patch representation.
Subsequently, all canonical patches are encoded and decoded by the AE, and the non-canonical form is returned (de-canonicalization).

Our contributions are summarized as follows:
\begin{itemize}\setlength{\parskip}{-2mm}
\item We extended MMES to MMQS by introducing equivalence classes and quotient spaces to image patches.
\item We established a non-trivial learning algorithm for MMQS and its application to several image reconstruction tasks, such as image inpainting, deblurring, super-resolution, and denoising.
\item We employed rotation-flip equivalence relations in MMQS and demonstrated the effectiveness of MMQS for image modeling in computational experiments.
\end{itemize}

\section{Proposed method}
In this section, we propose a new method for image reconstruction from corrupted images.

\subsection{Patch extraction from an image}
Let us consider $\bm X \in \bbR{I \times J}$ as a matrix that represents a single image of size $I \times J$. A patch of size $\sqrt{p} \times \sqrt{p}$ in $\bm X$ is represented as a vector $\bm x \in \bbR{p}$.
We define a set $\mathcal{X}$ as follow:
\begin{align}
  \mathcal{X} := \{ \bm x_1, \bm x_2, ..., \bm x_T \}, \label{eq:patch-extraction}
\end{align}
where each $\bm x_t$ is a patch extracted from $\bm X$, and $T$ is the number of extracted patches.

Practically, the extracted patches can be represented as a matrix as follows:
\begin{align}
  \mathcal{H}(\bm X) := [\bm x_1, ..., \bm x_T] \in \bbR{p \times T}, 
\end{align}
where the function $\mathcal{H}: \bbR{I \times J} \rightarrow \bbR{p \times T}$ represents the patch extraction process.
It can also be implemented by a convolutional layer using $p$ one-hot kernels.
The convolutional layer is usually included in typical deep learning platforms, such as \texttt{TensorFlow} and \texttt{PyTorch}.
One-hot kernels can be easily generated by reshaping the $(p \times p)$ identity matrix $\bm I$ into a $(\sqrt{p} \times \sqrt{p} \times p)$ tensor $\bm{\mathcal{I}}$.
Each frontal slice $\bm{\mathcal{I}}_{:,:,i} \in \bbR{\sqrt{p} \times \sqrt{p}}$ is an independent one-hot kernel.
Thus, we have $\mathcal{H}(\bm X) = \text{reshape}(\text{conv}(\bm X, \ten{I}))$, where conv($\cdot$,$\cdot$) is a function that represents a convolutional layer, and reshape($\cdot$) is a function to reshape a tensor into a matrix.

\subsection{Patch aggregation for image reconstruction}
Next, we define the process of patch aggregation for a single image reconstruction.
Patch aggregation can be simply thought of as the pseudo inverse of patch extraction.
Thus, patch aggregation is a function $\mathcal{H}^\dagger: \bbR{p \times T} \rightarrow \bbR{I \times J}$, and it satisfies
\begin{align}
  \mathcal{H}^\dagger \mathcal{H}(\bm X) = \bm X, \label{eq:pseudo-inverse-H}
\end{align}
for any $\bm X \in \bbR{I \times J}$.
In practice, this operation returns $T$ patches to their original locations in a single image. 
The value of the pixel where multiple patches overlap is calculated by its average.

Patch aggregation can be implemented by using a transposed convolutional layer.
Let $\bm 1$ be an all-one matrix of size $I \times J$, and calculate the following matrix in advance:
\begin{align}
  \bm D := \text{trconv}(\text{conv}(\bm 1, \ten{I}), \ten{I}) \in \bbR{I \times J},
\end{align}
where trconv($\cdot,\cdot$) represents a transposed convolutional layer.
Each entry of the matrix $\bm D$ is the number of overlapping patches during image reconstruction.
Using the matrix $\bm D$, the process of patch aggregation can be written as $\mathcal{H}^\dagger(\cdot) = \text{trconv}(\text{reshape}^{-1}(\cdot), \ten{I}) \oslash \bm D$, where $\text{reshape}^{-1}(\cdot)$ is an inverse function of $\text{reshape}(\cdot)$ and $\oslash$ is an element-wise division operation.

\subsection{Auto-encoder and manifold learning}
The AE plays the most important role in the proposed method.
We employ the denoising AE \cite{vincent2008extracting}.
The main function of the denoising AE is manifold learning, and it is expected that the learned manifold will have low dimensionality in order to be robust to additive noise.
It is also possible to explicitly enforce the low dimensionality of the manifold by setting the dimensionality of the intermediate layer to be small.

First, the denoising AE can be constructed by the following optimization problem:
\begin{align}
\mathop{\text{minimize}}_{\Theta} \mathbb{E}_{\bm x, \widetilde{\bm x} } \left[ || \bm x - \mathcal{A}_{\Theta}(\widetilde{\bm x}) ||_2^2 \right], \label{eq:DAE}
\end{align}
where $\bm x \in \mathcal{X} \subset \bbR{p}$ is a patch image of size $\sqrt{p} \times \sqrt{p}$ sampled from a set $\mathcal{X}$, and $\widetilde{\bm x} \sim N(\bm x, \bm I \sigma^2)$ is a noisy patch image corrupted by a zero-mean white Gaussian noise with variance $\bm I\sigma^2$.
Solving \eqref{eq:DAE} can be regarded as seeking a compact manifold $\mathcal{M}_\Theta$ such that $\mathcal{X} \subset \mathcal{M}_\Theta \subset \mathbb{R}^p$.

\begin{figure}[t]
  \centering
  \includegraphics[width=0.6\linewidth]{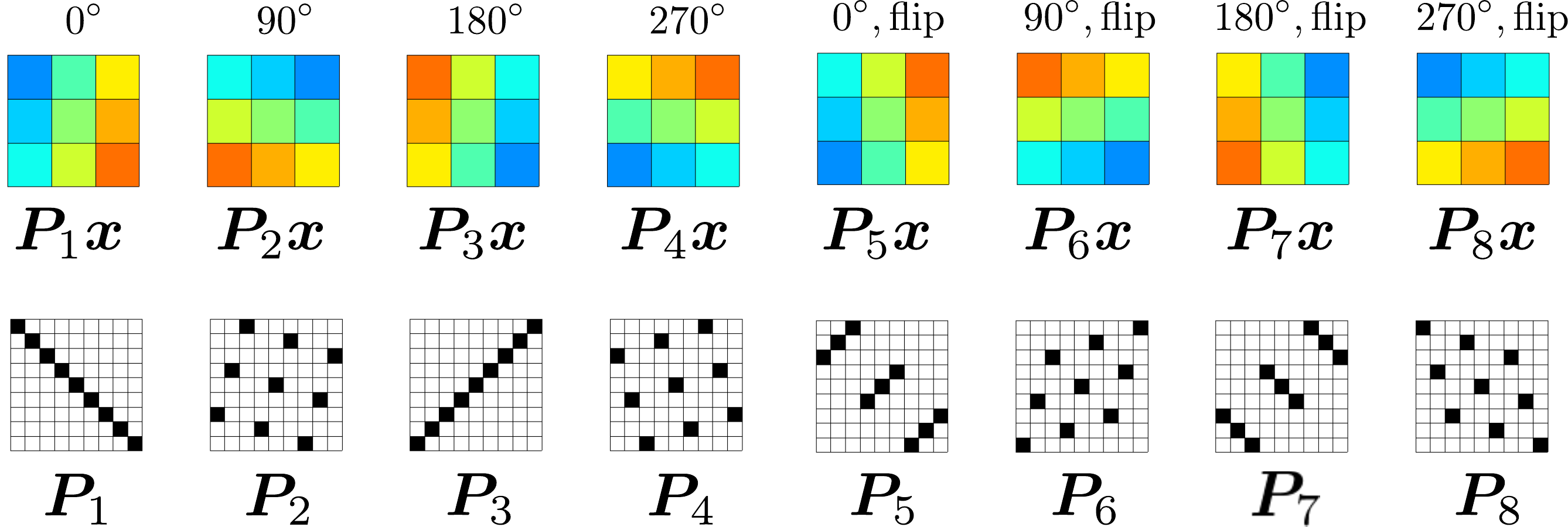}
\caption{Examples of eight operations for rotation and flip and their corresponding action matrices. A case of $p=9$. The second, third, and fourth action matrices ($\bm P_2$, $\bm P_3$, and $\bm P_4$) perform clockwise rotation operations of $90^\circ$, $180^\circ$, and $270^\circ$, respectively. The fifth action matrix ($\bm P_5$) performs a flip operation along the vertical direction. We have $\bm P_6 = \bm P_5 \bm P_2$, $\bm P_7 = \bm P_5 \bm P_3$, and $\bm P_8 = \bm P_5 \bm P_4$.}\label{fig:rotation-flip}
\end{figure}

\subsection{Manifold learning in quotient space}
Here, we introduce a quotient space based on equivalence relations.
First, we define a set of action matrices as follow:
\begin{align}
    \mathcal{R} := \{\bm P_1, ..., \bm P_K\},
\end{align}
where each $\bm P_k \in \bbR{p \times p}$ must be regular (\ie $\bm P_k^{-1}$ exists).
In this study, we employ eight action matrices for rotation and flip (see \cref{fig:rotation-flip}).
The equivalence class of $\bm x \in \mathcal{X}$ is given by
$[\bm x] := \{\bm P \bm x | \bm P \in \mathcal{R} \}$.
Then, the quotient set $\mathcal{X}/\mathcal{R} := \{ [\bm x] \ | \  \bm x \in \mathcal{X} \}$ can be defined, and we want to seek a compact manifold $\mathcal{Q}$ such that $\mathcal{X}/\mathcal{R} \subset \mathcal{Q} \subset \bbR{p}/\mathcal{R}$.
However, $\mathcal{Q}$ is a set of equivalence classes, and it is difficult to represent it using a parametric model. Therefore, we consider embedding $\mathcal{X}/\mathcal{R}$ into a vector space by canonicalization; a conventional manifold learning is applied to it.

In practice, we consider the following problem:
\begin{align}
\mathop{\text{minimize}}_{\Theta} \mathbb{E}_{\bm x, \widetilde{\bm x}} \left[ \min_{\bm P \in \mathcal{R} } || \bm P \bm x - \mathcal{A}_{\Theta}(\bm P \widetilde{\bm x}) ||_2^2 \right], \label{eq:DAE-RF}
\end{align}
where $\bm P$ represents a regular matrix that performs the canonicalization of patch $\bm x$.
Note that the best $\bm P$ is different for each $\bm x \in \mathcal{X}$, but $\bm P^{(i)} \bm x_i = \bm P^{(j)} \bm x_j$ is satisfied when $[\bm x_i] = [\bm x_j]$ because of the $\min$ operation.
In other words, all patches with equivalence relations are canonicalized as the same patch by the min operation.
We call it a canonical auto-encoder (CAE) in this paper.
Note that our formulation \eqref{eq:DAE-RF} is a generalization of \eqref{eq:DAE}.
When we set $\mathcal{R} = \{ \bm I \}$, our model \eqref{eq:DAE-RF} is reduced to the normal denoising AE \eqref{eq:DAE}.

Note that the proposed CAE is very different from the normal AE with data augmentation.
A normal AE with data augmentation learns common features for all \{$\bm P_1 \bm x$, ..., $\bm P_K \bm x\}$; however, the proposed CAE adaptively selects only one action matrix from $K$ candidates for each $\bm x$.  In other words, data augmentation expands the data distribution; conversely, our model is expected to shrink the data distribution.

\subsection{Patch manifold learning from a single image}\label{sec:manifold-learning}
Let us assume that a single image $\bm X$ is given; we consider learning a patch-based AE from $\bm X$ in this section.
In other words, the patch extraction from a single image \eqref{eq:patch-extraction} and manifold learning with CAE \eqref{eq:DAE-RF} are combined as follows:
\begin{align}
\mathop{\text{minimize}}_{\Theta} \mathbb{E}_{\bm n} \sum_{t=1}^T \left[ \min_{\bm P^{(t)} } || \bm P^{(t)} \bm x_t - \mathcal{A}_{\Theta}(\bm P^{(t)} \bm x_t + \bm n ) ||_2^2 \right], \label{eq:opt-DAE-RF}
\end{align}
where $\bm n \sim N(\bm 0, \bm I \sigma^2)$ is the white Gaussian noise.
The optimization parameters is not only $\Theta$ of AE, but also $\bm P^{(t)} \in \mathcal{R}$ for each $\bm x_t \in \mathcal{X}$.
These are summarized as $\{\Theta, \bm P^{(1)}, ..., \bm P^{(T)}\}$.

Especially for the action matrices $ \{ \bm P^{(t)} \}_{t=1}^T$, Eq.~\eqref{eq:opt-DAE-RF} is a combinatorial optimization problem with $K^T$ cases.
Therefore, it is difficult to find a global solution to this optimization problem.
However, if $\Theta$ is fixed, each action matrix $\bm P^{(t)}$ can be regarded as an independent optimization parameter, and the optimization problem can be separated into individual discrete optimization problems for $T$ action matrices. Furthermore, if all the action matrices are fixed, the cost function can be reduced by updating $\Theta$ using the gradient descent method. Thus, a local solution can be obtained by applying alternating optimizations on $\Theta$ and action matrices $\{ \bm P^{(t)} \}_{t=1}^T$.

Finally, the proposed alternating optimization algorithm for learning CAE is as follows. For each $t$, the $t$-th action matrix is updated by
\begin{align}
    \bm P^{(t)} \leftarrow \mathop{\text{argmin}}_{\bm P \in \mathcal{R} } || \bm P \bm x_t - \mathcal{A}_\Theta (\bm P \bm x_t) ||_2^2. \label{eq:update-P}
\end{align}
For a cost function defined by 
\begin{align}
  \mathcal{L}(\Theta) := \sum_{t=1}^T || \bm P^{(t)} \bm x_t - \mathcal{A}_{\Theta}(\bm P^{(t)} \bm x_t + \bm n_t ) ||_2^2, \label{opt:DAE-loss}
\end{align}
where each $\bm n_t$ is randomly generated by a Gaussian distribution $N(\bm 0, \bm I \sigma^2)$, and the optimization parameter $\Theta$ is updated using the gradient descent method:
\begin{align}
    \Theta_{\text{new}} \leftarrow \Theta_{\text{old}} - \eta \frac{\partial \mathcal{L}}{\partial \Theta}(\Theta_\text{old}),
\end{align}
where $\eta > 0$ is the learning rate.
In practice, various gradient-descent-based algorithms such as Adam \cite{kingma2014adam} are applicable.

\subsection{Concept of image model}\label{sec:image-model}
In contrast, \Cref{sec:manifold-learning} explains a method for learning a patch-based AE from a given single image $\bm X$. Here, we consider reconstructing an image $\bm X$ using an image model defined by a patch manifold.
Let us assume that a CAE $\mathcal{A}_{\Theta}$ is given and that it can be considered as an image model:
\begin{align}
  \mathcal{S}_\Theta := \{ \bm X \ | \ \forall \bm x \in \mathcal{X}, \exists k \in \mathcal{K}, || \bm P_k \bm x - \mathcal{A}_{\Theta} \bm P_k \bm x ||_2^2 \leq \epsilon \},\label{eq:image_model}
\end{align}
where $\mathcal{K} := \{1, 2, ..., K \}$, and $\epsilon > 0$ is a small value.
We consider an image model represented by the set $\mathcal{S}_\Theta$ of all possible images $\bm X$ that satisfy the above condition.
This means that all patches of the image $\bm X$ included in $\mathcal{S}_\Theta$ are around the manifold represented by AE $\mathcal{A}_{\Theta}$.

An image reconstruction problem using the given model $\mathcal{S}_\Theta$ can be formulated as follows:
\begin{align}
    \widehat{\bm X} = \mathop{\text{argmin}}_{\bm X} || \bm Y - \mathcal{F}(\bm X)  ||_F^2, \text{ s.t. } \bm X \in \mathcal{S}_\Theta,\label{eq:image_model_2}
\end{align}
where $\bm Y$ is an observed (usually corrupted) image, and $\mathcal{F}$ is a linear operator that represents the observation model.
For example, we set $\mathcal{F}$ as an identity mapping for the denoising task, a masking operator for the inpainting task, a blurring operator for the deblurring task, and a downsampling operator for the super-resolution task.

Combining \eqref{eq:image_model} and \eqref{eq:image_model_2}, the following optimization problem is obtained:
\begin{align}
    \mathop{\text{minimize}}_{\bm X} & \ || \bm Y - \mathcal{F}(\mathcal{H}^\dagger \mathcal{P}^{-1} \mathcal{A}_{\Theta} \mathcal{P} \mathcal{H} (\bm X))  ||_F^2 \notag \\ 
     & \ \ \  + \lambda || \mathcal{P} \mathcal{H} (\bm X) - \mathcal{A}_{\Theta} \mathcal{P} \mathcal{H} (\bm X) ||_F^2, \label{eq:image-reconstruction-problem} \\
     \text{s.t. } & \bm P^{(t)} = \mathop{ \text{argmin} }_{ \bm P \in \mathcal{R} } || \bm P \bm x_t - \mathcal{A}_{ \Theta } \bm P \bm x_t ||_2^2, \notag
\end{align}
where $\mathcal{P}([\bm x_1, ..., \bm x_T]) := [\bm P^{(1)}\bm x_1, ..., \bm P^{(T)}\bm x_T]$, and $\mathcal{P}^{-1}([\bm x_1, ..., \bm x_T]) := [\bm P^{(1)\top}\bm x_1, ..., \bm P^{(T)\top}\bm x_T]$.
This optimization problem can be solved by alternating the optimization with respect to $\bm X$ and $\bm P^{(t)} (\forall t)$.
Finally, the reconstructed image can be obtained by $\widehat{\bm X} = \mathcal{H}^\dagger \mathcal{P}^{-1}_\text{last} \mathcal{A}_\Theta \mathcal{P}_{\text{last}} \mathcal{H} (\bm X_\text{last})$.

\subsection{Self-supervised image reconstruction from a corrupted image}
In \Cref{sec:manifold-learning}, we considered learning CAE $\mathcal{A}_\Theta$ from a given clean image $\bm X$.
In \Cref{sec:image-model}, we reconstructed an image $\bm X$ from a corrupted image $\bm Y$ by using a given CAE $\mathcal{A}_\Theta$.
In this section, we consider a case in which both a clean image $\bm X$ and a CAE $\mathcal{A}_\Theta$ are not given, but only a corrupted image $\bm Y$ is given and its observation model $\mathcal{F}$ is known.

This study aims to simultaneously learn a CAE $\mathcal{A}_\Theta$ and reconstruct an image $\bm X$ from the corrupted image $\bm Y$.
We combine the techniques explained in \Cref{sec:manifold-learning,sec:image-model} and propose a new self-supervised image reconstruction model as follows:
\begin{align}
        \mathop{\text{minimize}}_{\bm X, \Theta} & \ \mathcal{L}_\lambda(\bm X, \Theta) := \mathcal{L}_\text{rec}(\bm X, \Theta) + \lambda \mathcal{L}_\text{CAE}(\bm X, \Theta), \notag \\
     \text{s.t. } & \bm P^{(t)} = \mathop{ \text{argmin} }_{ \bm P \in \mathcal{R} } || \bm P \bm x_t - \mathcal{A}_{\Theta} \bm P \bm x_t ||_2^2, \label{eq:image-reconstruction-problem-self-supervised}
\end{align}
where 
\begin{align}
    &\mathcal{L}_\text{rec}(\bm X, \Theta) := || \bm Y - \mathcal{F}(\mathcal{H}^\dagger \mathcal{P}^{-1} \mathcal{A}_\Theta \mathcal{P} \mathcal{H} (\bm X))  ||_F^2, \\ 
    &\mathcal{L}_\text{CAE}(\bm X, \Theta) := || \mathcal{P} \mathcal{H} (\bm X) - \mathcal{A}_\Theta \mathcal{N}_\sigma \mathcal{P} \mathcal{H} (\bm X) ||_F^2,
\end{align}
and $\mathcal{N}_\sigma([\bm x_1, ..., \bm x_T]) := [\bm x_1 + \bm n_1, ..., \bm x_T + \bm n_T]$ is an operator for adding Gaussian noise $\bm n_t \sim N(\bm 0, \sigma^2 \bm I)$ for all $t \in \{1, ..., T\}$.

Similar to \eqref{eq:DAE-RF} and \eqref{eq:image-reconstruction-problem}, we solve \eqref{eq:image-reconstruction-problem-self-supervised} by alternating the optimization with respect to $(\bm X, \Theta)$ and $\{\bm P^{(t)}\}_{t=1}^T$.
For each $t$, the $t$-th action matrix is updated using \eqref{eq:update-P}.
We update $(\bm X, \Theta)$ by
\begin{align}
    (\bm X_\text{new}, \Theta_{\text{new}}) \leftarrow (\bm X_\text{old}, \Theta_{\text{old}}) - \eta \frac{\partial \mathcal{L}_\lambda}{\partial (\bm X, \Theta)}(\bm X_\text{old}, \Theta_\text{old}).
\end{align}
Note that the minimization of the cost function $\mathcal{L}_\lambda$ with a balance parameter $\lambda > 0$ is somewhat sensitive, with a value of $\lambda$. When $\lambda$ is too large, it is difficult for the reconstructed image to fit the observed image, and optimization sometimes fails. Following the optimization strategy in \cite{yokota2020manifold}, we also adjust $\lambda$ based on the balance between $\mathcal{L}_\text{rec}$ and $\mathcal{L}_\text{CAE}$.

\section{Related works}
Studies on unsupervised (or self-supervised) image modeling using neural network structures have been actively conducted in recent years \cite{ulyanov2018deep,ulyanov2020deep,heckel2018deep,shaham2019singan,shocher2019ingan,lehtinen2018noise2noise,krull2019noise2void,batson2019noise2self,xu2020noisy,cha2019gan2gan,anirudh2021generative,yokota2020manifold}.
Among these, image modeling methods target only the image denoising task.
In contrast, in this study, denoising and many other image reconstruction tasks based on arbitrary linear observation models such as inpainting, deblurring, and super-resolution can be solved.
The most related studies are DIP \cite{ulyanov2018deep, ulyanov2020deep} and MMES \cite{yokota2020manifold}.
In particular, MMQS can be characterized as a generalization of MMES.

In addition, the approach using quotient spaces in computer vision has been studied for many years.
In \cite{riklin1999quotient, shashua2001quotient, wang2004generalized}, the concept of a quotient image that normalizes the lighting conditions to improve the accuracy of face recognition was proposed and studied.
In \cite{tron2014quotient}, the quotient space was introduced as a representation of the essential matrices in the study of multiple-view geometry.
In \cite{mehr2018manifold}, a quotient auto-encoder (QAE) was proposed for point cloud processing.
The QAE \cite{mehr2018manifold} and the proposed CAE share the same concept, but their structures are slightly different.
The QAE uses an orbit pooling layer that performs max pooling for all latent representations in the same equivalence class instead of the canonicalization layer in the CAE.  Thus, latent (canonical) representation in QAE is a mixture of all latent representations; in contrast, the proposed CAE explicitly selects only one representation from them.

The idea of utilizing rotated and flipped images as an equivalence class can be interpreted as adopting a graph representation of the image
if the image is represented by each pixel as a node, the adjacency of the pixels as edges, and all the rotated and flipped images are graph isomorphisms.
The measuring similarity between images in a graph representation is generalized as a graph matching problem, such as the graph edit distance \cite{bunke1997relation,bunke1998graph,messmer1998new}. Such an approach naturally introduces a quotient space formulation that essentially treats isomorphic graph data as equivalence classes \cite{chowdhury2020gromov, guo2020representations, guo2021quotient}.

Non-local similarity filtering methods that describe the relationships between pixels using the distance in a feature space that does not rely on rotation and flip have also been studied \cite{lou2009nonlocal, ji2009moment, grewenig2011rotationally}.

Note that the proposed approach is different from the minimization of contrastive loss with data augmentation \cite{hadsell2006dimensionality,karianakis2013learning,dosovitskiy2014discriminative,oord2018representation,bachman2019learning,chen2020simple}.
Since our approach is {\em separation} for variant manipulation rather than {\em deletion} (as in contrasting learning); the decodability is preserved, even after encoding.

\begin{figure}[t]
  \centering
      \includegraphics[width=0.49\textwidth]{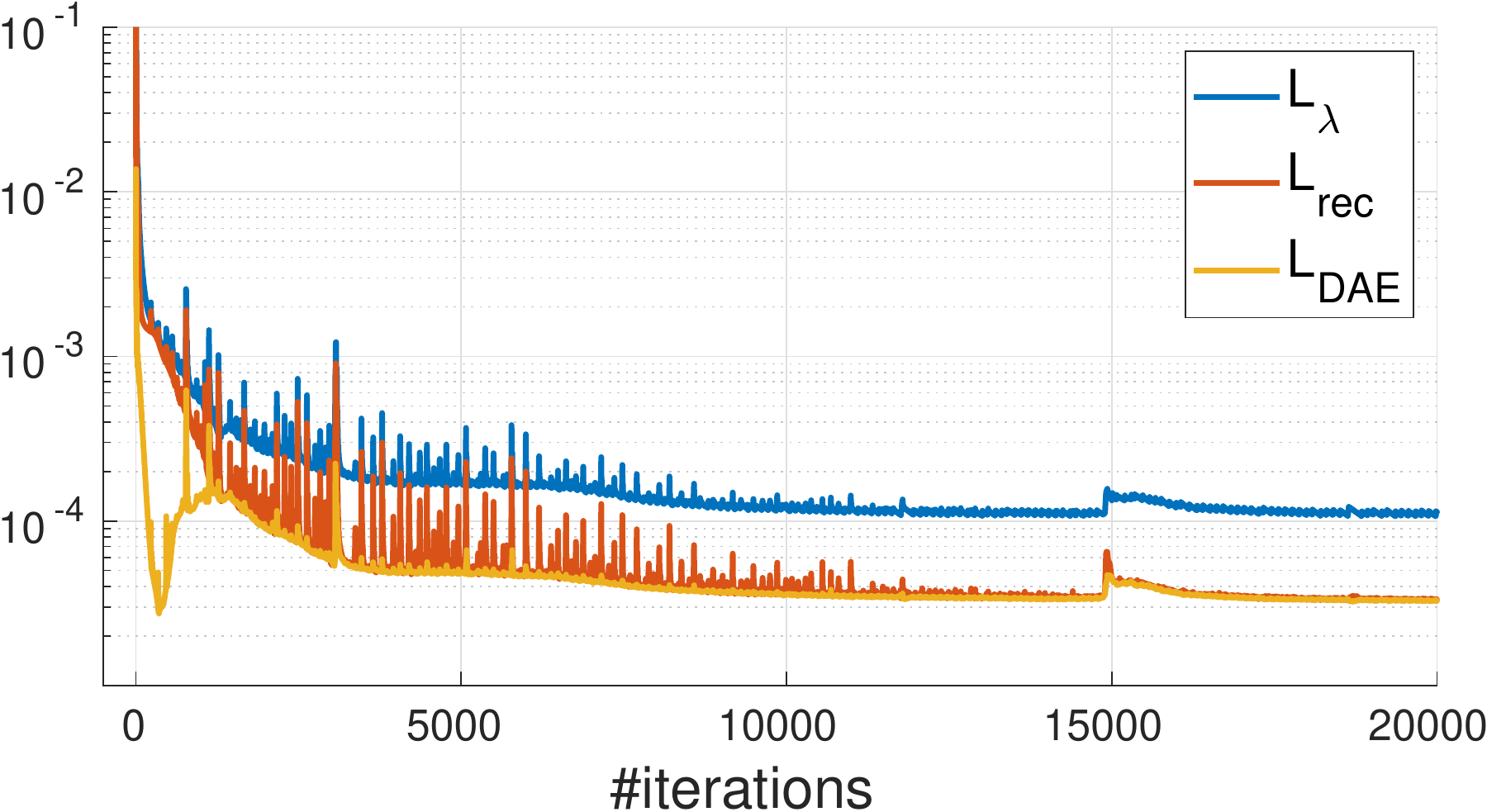}
    \caption{Optimization behavior.}
    \label{fig:opt_behav}
\end{figure}
\begin{figure}
  \centering
\includegraphics[width=0.7\textwidth]{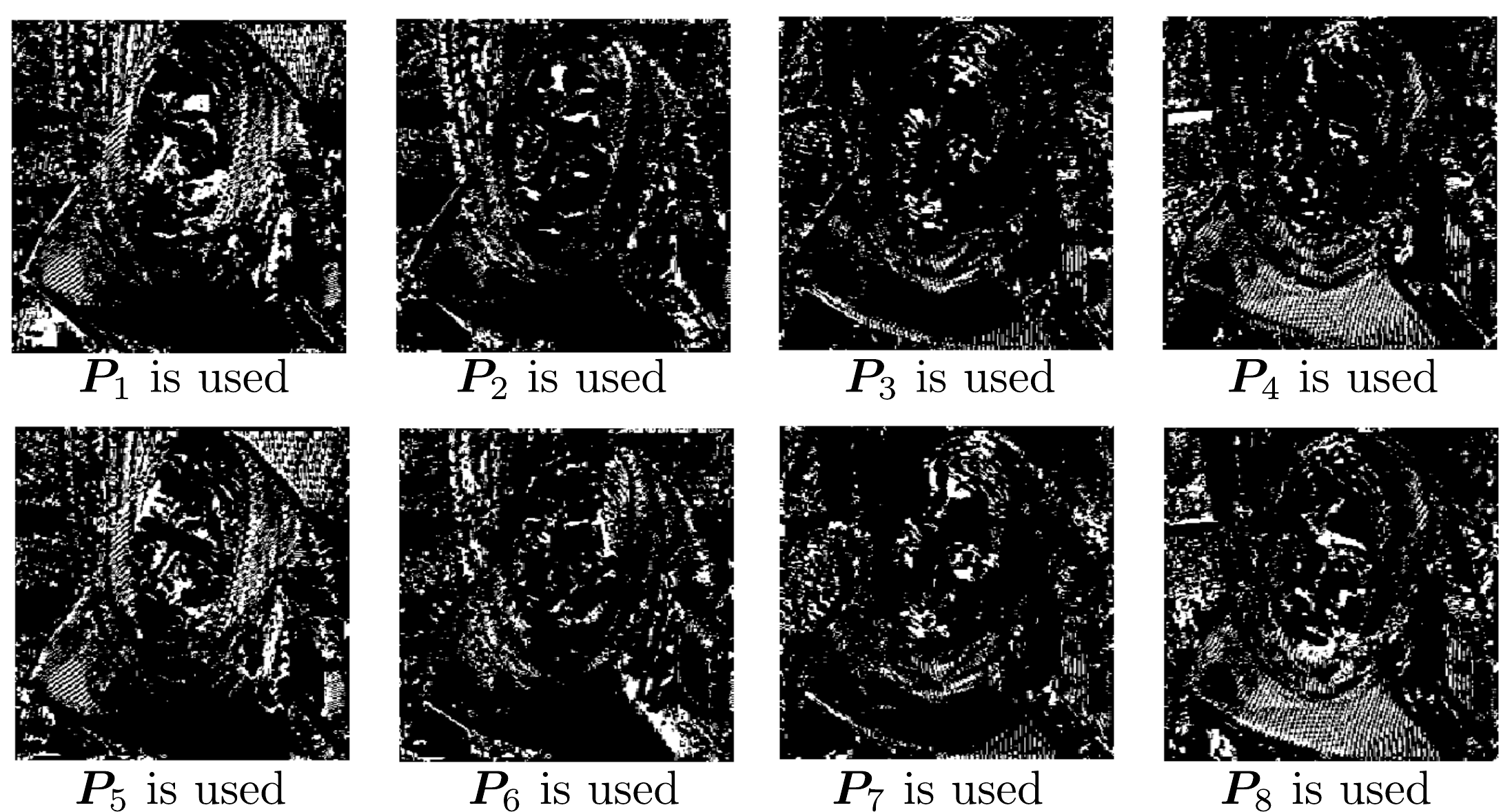}
    \caption{Visualization of the selected action matrices.}
    \label{fig:P_labels}
\end{figure}

\begin{figure}[t]
  \centering
   \includegraphics[width=0.99\linewidth]{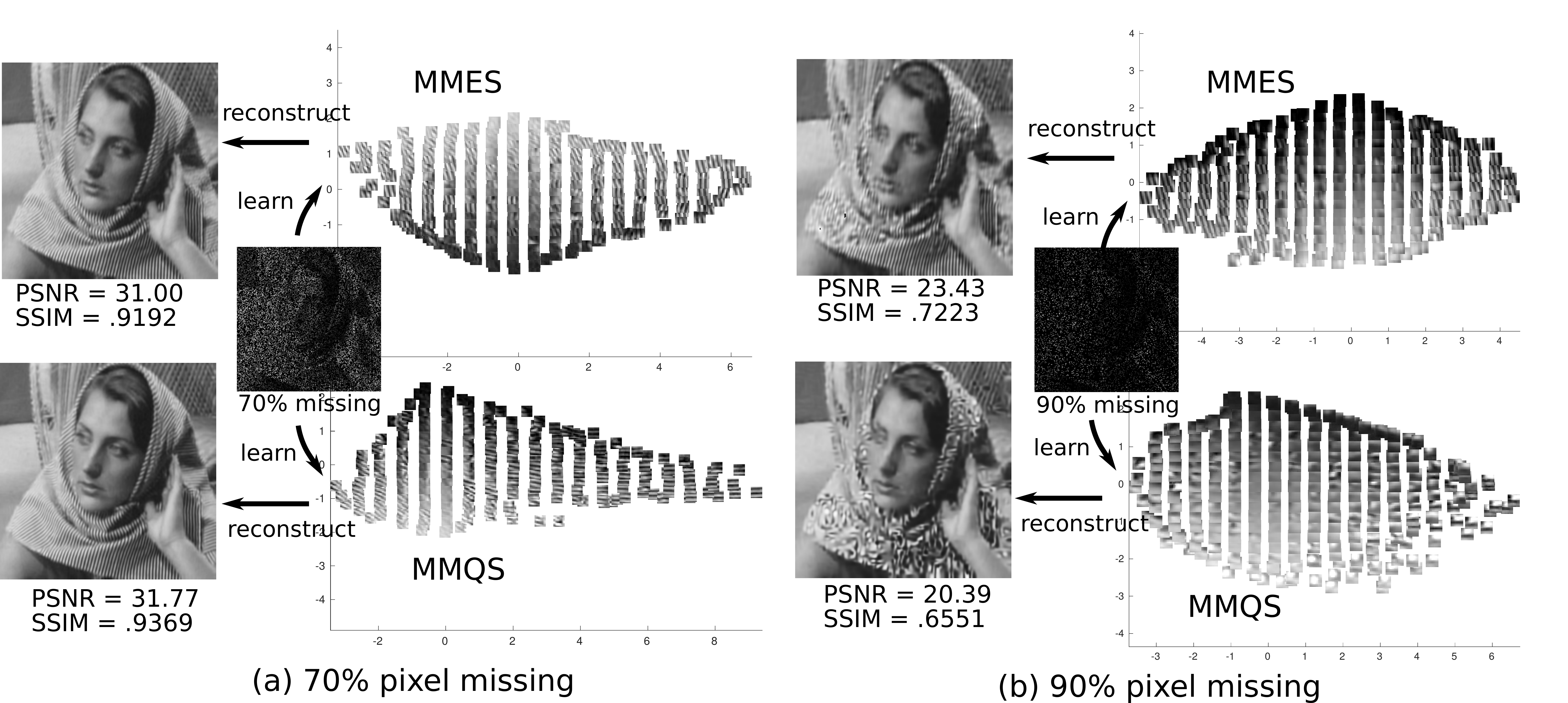}
   \caption{Results of the inpainting task and visualization of canonical patches on the learned manifold.}
   \label{fig:manifold}
   \vspace{3mm}
%\end{figure}
%\begin{figure}[t]
  \centering
      \includegraphics[width=0.6\textwidth]{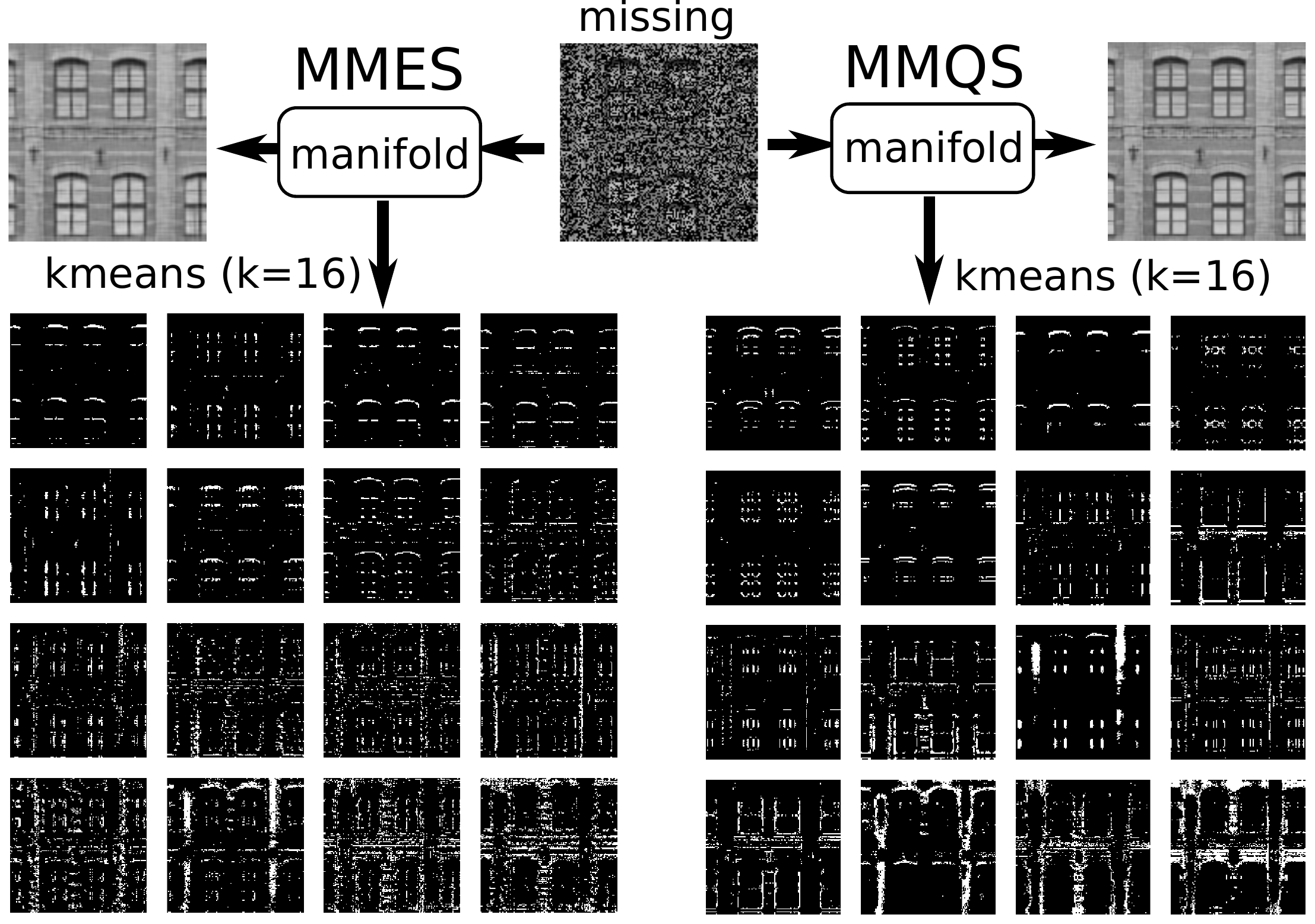}
    \caption{Structure analysis of image patches.}
    \label{fig:opt_kmeans}
\end{figure}

\section{Experiments}

\subsection{Standard behaviors with visualization of patch-manifolds}\label{sec:opt}
In this section, we show the standard behavior of the proposed MMQS and compare it to that of MMES.
A grayscale image named `Barbara' with a size of 256 × 256 was used for this experiment.
70\% or 90\% of pixels were randomly removed from `Barbara', thus creating two incomplete images.
From the two gray-scaled images with missing pixels, we learned the patch manifolds and reconstructed the images using MMES and MMQS; the results were subsequently compared.
Hyperparameters were set to $\tau$ = 9 and $\sigma$ = 0.05.
The structure of the AE was a multi-layered perceptron (MLP) using the Leaky ReLu as an activation function, and the size of the intermediate layer was (81, 81, 10, 81, 81).
The Adam optimizer was employed for weight update while adjusting $\lambda$ such that $\mathcal{L}_\text{rec}$ and $\mathcal{L}_\text{CAE}$ were minimized in a well-balanced manner.

\cref{fig:opt_behav} shows the optimization behavior when MMQS reconstructed the image from 70\% pixel missing.
In the early stages of optimization, the CAE loss decreases rapidly and the reconstruction loss decreases relatively slowly.
At this stage, the patch AE may still be trivial and does not reflect the input image.
Subsequently, to apply the local patterns of the reconstructed image to the input image, the CAE loss temporarily increases, and manifold learning essentially begins from here.
Finally, the overall optimization converges as the learning of the patch manifold converges.

\cref{fig:P_labels} shows a visualization of what the permutation matrix for the canonicalization of each patch looks like as a result of the optimization.
It can be seen that the same permutation matrix is selected for similar patches.

\cref{fig:manifold}(a) shows the visualization of the patch manifold learned from the image where 70\% of the pixels are missing.
The dimension of the manifold was 10, and the 9 × 9 canonical patches were plotted in 2D space obtained via principal component analysis.
In the patch manifold of MMES, the stripe patterns oriented in various directions are mixed, whereas, in the patch manifold of MMQS, the stripe patterns are oriented in the horizontal direction.
In addition, comparing the reconstructed images, it can be seen that MMQS outperforms MMES.
In particular, MMES fails to restore the horizontal stripes of the scarf located at Barbara's occipital part, whereas MMQS is able to restore it cleanly.

\cref{fig:manifold}(b) shows the results for the case where 90\% of the pixels are missing.
In the patch manifold of the MMES, the vertical stripe patterns are dominant, and the reconstruction result is also affected. In particular, stripe pattern artifacts are present on the face.
However, in MMQS, the stripe pattern is not learned, and the reconstruction result is characterized by the mottled pattern of the scarf facing in various directions. Unlike MMES, there are no stripe pattern artifacts on the face.
In this case, the recovery accuracy of MMQS for true images is inferior to that of MMES, but this is not a problem for our motivation.
MMQS is a softer image model than MMES because there are no restrictions on the orientation of the patches in the reconstructed image. It is expected that image reconstruction with a strongly ill-posed setting, such as 90\% of the data missing, will be disadvantageous.
Conversely, this result can be considered as evidence that the concept of rotational-flip-equivalence class, which is the key idea in this study, has been realized.

\subsection{Analysis of similar patch subsets}
Next, we show the results of the $k$-means clustering of intermediate representations in an AE trained from an incomplete image where 50\% of the pixels are missing by MMES and MMQS.
We set the size of the intermediate layers as (81, 81, 10, 81, 81) and extracted the central 10-dimensional representations.
$k$-means clustering with $k=16$ was applied to both the MMES and MMQS cases.

\cref{fig:opt_kmeans} shows the results.
Clusters of similar patches based on the MMES and MMQS criteria are visualized.
MMQS seems to have a clearer structure than MMES.
In particular, MMQS contains patch clusters that surround the window frames in the image.

\begin{figure}[t]
  \centering
   \includegraphics[width=0.99\linewidth]{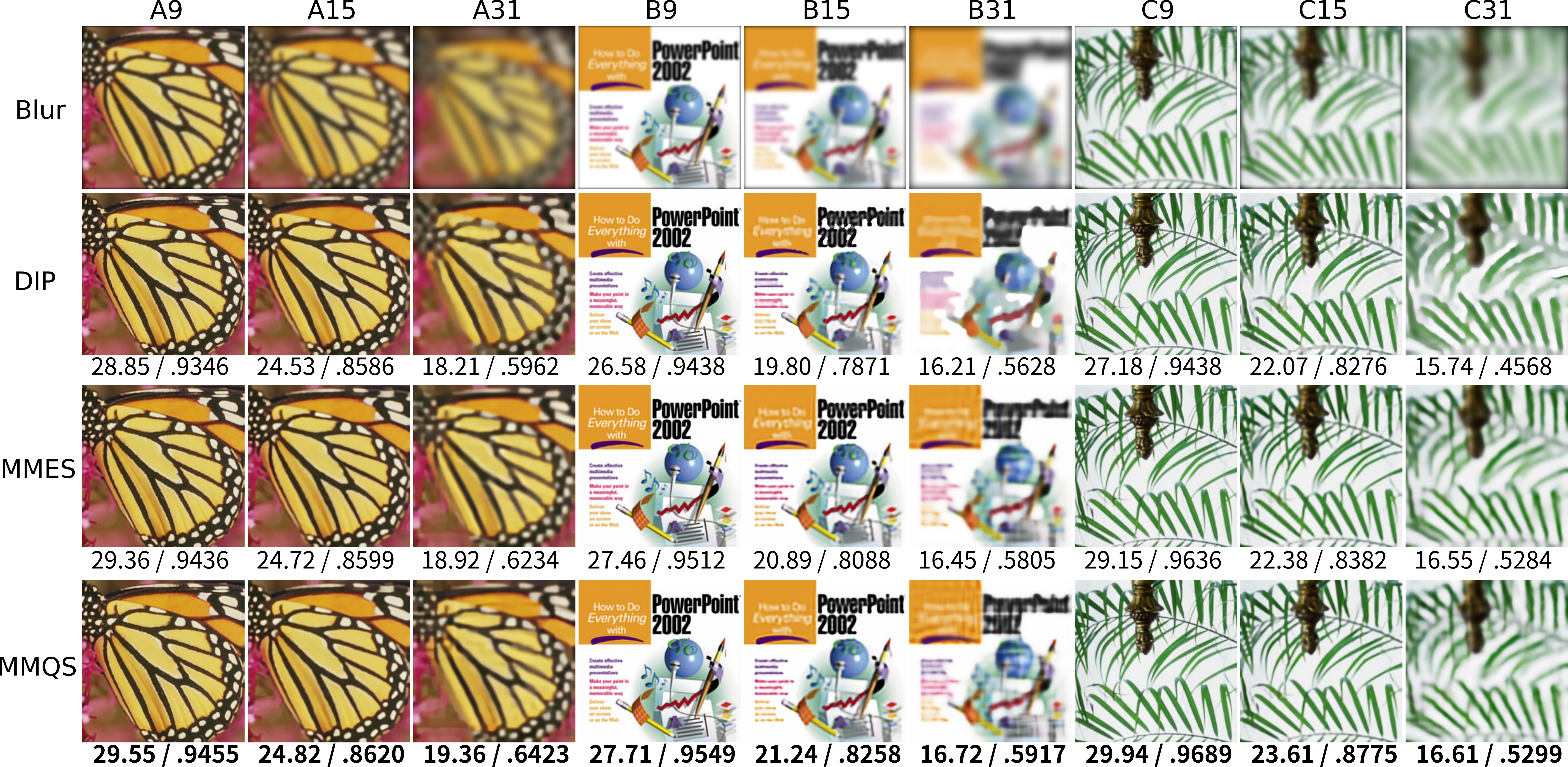}
   \caption{Results of image deblurring.}
   \label{fig:deblurring}
\end{figure}

\begin{figure}[t]
  \centering
   \includegraphics[width=0.99\linewidth]{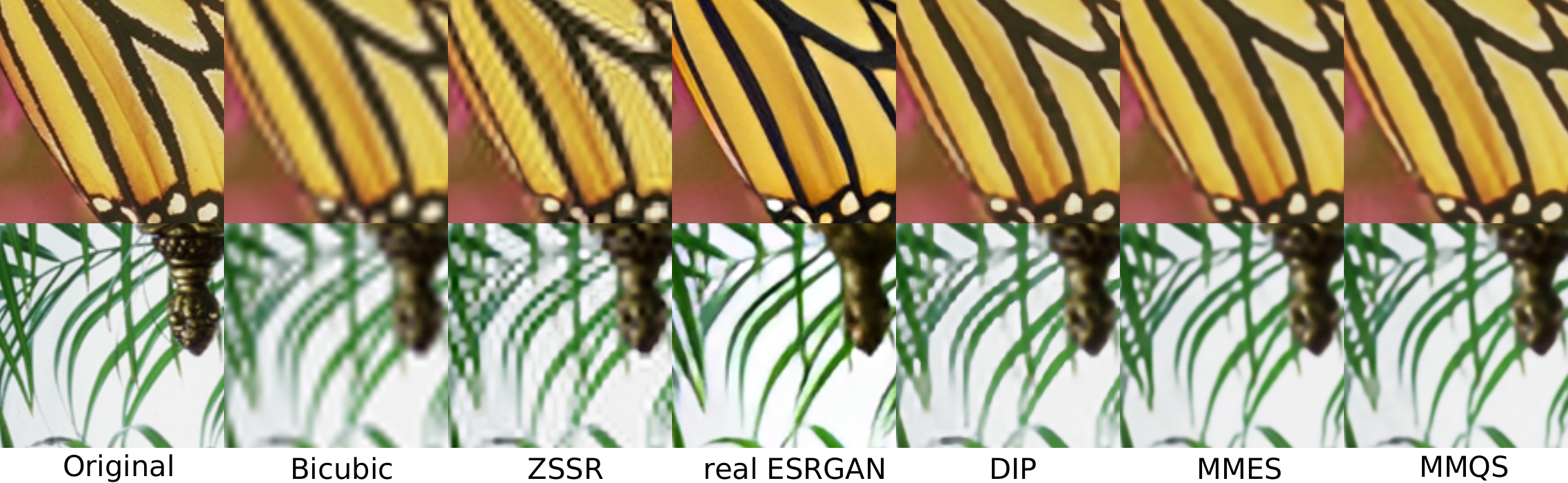}
   \caption{Results of image super-resolution.}
   \label{fig:super-resolution}
   %\vspace{5mm}
\end{figure}

\subsection{Color image deblurring}
%In this section, we show the experimental results of color image deblurring.
Three color images of sizes 256 × 256 × 3 were prepared and artificially blurred by Gaussian kernel functions with widths of 9, 15, and 31, respectively.
We assumed that the Gaussian kernel functions were known in advance, and reconstructed the images using DIP, MMES, and MMQS.
The settings of MMQS were $p = 16$ and $\sigma = 0.01$, and the sizes of the MLP intermediate layers were (512, 16, 512).

\cref{fig:deblurring} shows the result of color image deblurring.
The numerical value written below each image is PSNR / SSIM.
It can be observed that MMQS outperforms other methods for both PSNR and SSIM.

\subsection{Color image super-resolution}
%In this section, we present the experimental results of the color image super-resolution.
Four color images of sizes 256 × 256 × 3 and four color images of sizes 512 × 512 × 3 (8 in total) were prepared and downscaled to 1/4.
We assumed that the linear downscaling operator was known in advance, and upscaled reconstruction was performed using bicubic interpolation, ZSSR \cite{shocher2018zero}, real ESRGAN \cite{wang2021realesrgan}, DIP, MMES, and MMQS. Note that real ESRGAN is slightly different from other methods in that it is a method that learns an image prior from a large amount of image data. 
The settings of MMQS were $p = 36$ and $\sigma = 0.1$, and the sizes of the MLP intermediate layers were (288, 16, 288).

\cref{fig:super-resolution} shows the results of the reconstructed images, and \cref{tab:PSNR_SSIM} shows the evaluation results by PSNR and SSIM.
In Bicubic interpolation, the image are significantly blurred.
Ringing-like artifacts occur in the ZSSR.
The image reconstructed by real ESRGAN looks great; however, the contrast of the image is overemphasized. The results evaluated by PSNR and SSIM are considerably inferior to those of the other methods.
In DIP, the expression of the diagonal fine edges is not good, especially in the leaves; the tip of the leaf was zigzag.
It can be seen that MMES and MMQS are well represented by fine edges, and PSNR and SSIM are also superior to the other methods.

\begin{figure}[t]
  \centering
  \includegraphics[width=0.99\linewidth]{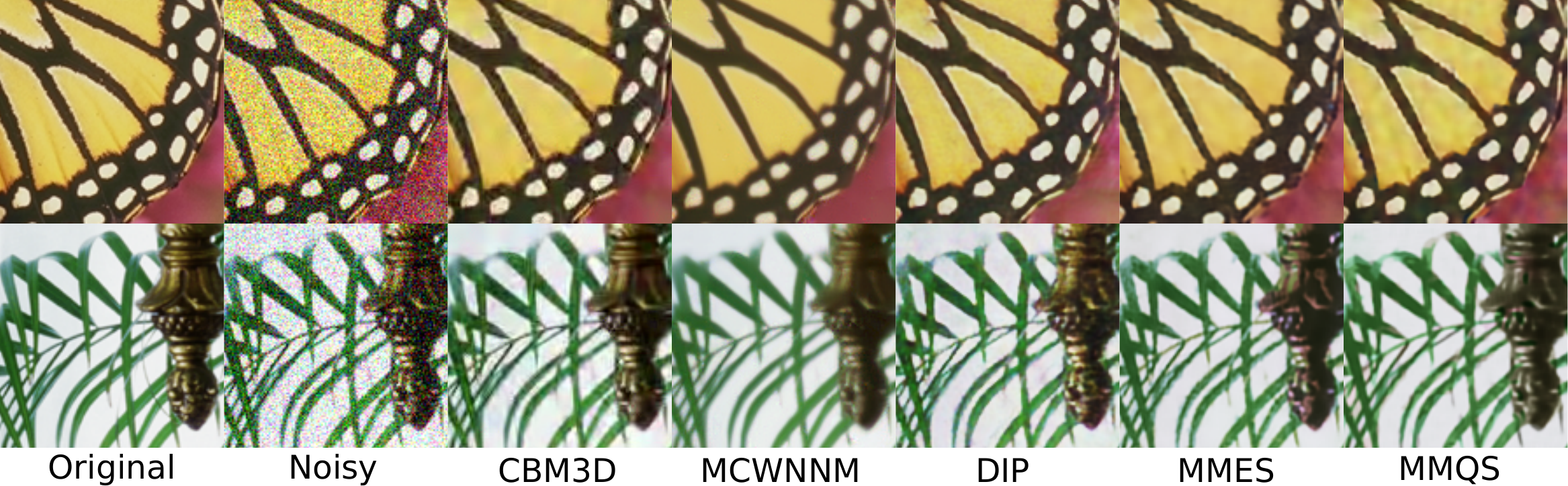}
   \caption{Results of image denoising.}
   \label{fig:denoising}
\end{figure}

\subsection{Color image denoising}
Seven images of size 256x256x3 with additive noise ($\sigma=40$) were prepared for this experiments.
We assumed noise variance is known, and reconstruct them by CBM3D \cite{dabov2007color}, MCWNNM \cite{xu2017multi}, DIP, MMES, and MMQS.
The settings of MMQS were $p = 36$ and $\sigma = 0.05$, and the sizes of the MLP intermediate layers were (288,36,288).
In optimization, we adjusted $\lambda$ to keep small CAE loss, and used early stopping like DIP.

\cref{fig:denoising} and \cref{tab:PSNR_SSIM} show the results.
The CBM3D and the proposed MMQS performed the top and second-top reconstruction.

\begin{table}
  \centering
  {\small
  \begin{tabular}{@{}l|ccc@{}}
    \toprule
    Methods         & Deblurring    & Super-resolution & Denoising \\
    \midrule
    MMQS (prop.)    & {\bf 23.28}/{\bf .7998}   & {\bf 26.66}/\underline{.8792}   &  \underline{27.03}/\underline{.8526} \\
    MMES \cite{yokota2020manifold}           & \underline{22.86}/\underline{.7886}   & \underline{26.59}/{\bf .8806}   &  26.93/.8441 \\
    DIP \cite{ulyanov2018deep}            & 22.15/.7680   & 26.18/.8705   &  25.81/.8198 \\
    Bicubic         & N.A.          & 23.96/.7952   & N.A.   \\
    ZSSR \cite{shocher2018zero}           & N.A.          & 24.84/.8331  & N.A.   \\  
    real ESRGAN \cite{wang2021realesrgan}     & N.A.          & 23.37/.8236  & N.A.   \\
    CBM3D \cite{dabov2007color}          & N.A.          & N.A.  & {\bf 27.65}/{\bf .8650}  \\
    MCWNNM  \cite{xu2017multi}        & N.A.          & N.A.  &  26.36/.8363 \\
    \bottomrule
  \end{tabular}
  }
  \caption{Quantitative evaluation of the color image reconstruction tasks. The best values are highlighted in bold, and the second-best values are underlined.}
  \label{tab:PSNR_SSIM}
\end{table}

\section{Limitations and conclusions}
In this study, we introduced equivalence classes based on a set of actions for the image space and proposed a novel approach, MMQS, to learn the manifold of the equivalence classes. As a concrete example of the action set, we adopted eight types of permutation matrices, namely, image rotation and flip, and conducted experiments.
The size of the action set could be increased, and various actions other than rotation and flip can also be employed.
However, there are generally the following limitations.
First, each action must be invertible.
Second, the encoding cost increases linearly with respect to the size of the action set. Third, the selection of actions is discrete and not parameterized with continuous variables. Originally, image rotation was an action with a continuous angular parameter, but this time it was discretized with a 90$^\circ$ interval.
Future works will include the improvement of scalability of the action set, the establishment of continuous parameterization of actions, improvement of the optimization algorithm, and applications to other tasks.

%%%%%%%%% REFERENCES
\bibliographystyle{plain}
%\bibliography{egbib}

\end{document}